\pdfoutput=1

\documentclass[11pt]{article}

\usepackage[preprint]{acl}
\usepackage{stfloats}

\usepackage{times}
\usepackage{latexsym}
\usepackage{todonotes}

\usepackage[T1]{fontenc}

\usepackage[utf8]{inputenc}

\usepackage{microtype}

\usepackage{inconsolata}

\usepackage{graphicx}

\usepackage{hyperref}
\usepackage{url}
\usepackage{multirow}

\title{Training and Inference Efficiency of Encoder-Decoder Speech Models}

\author{
    Piotr Żelasko,  Kunal Dhawan, Daniel Galvez, Krishna C. Puvvada, Ankita Pasad, \\
    {\bf Nithin Rao Koluguri, Ke Hu, Vitaly Lavrukhin, Jagadeesh Balam, Boris Ginsburg} \\
        NVIDIA, USA \\
        \texttt{pzelasko@nvidia.com}
        }

\begin{document}
\maketitle
\begin{abstract}
Attention encoder-decoder model architecture is the backbone of several recent top performing foundation speech models: Whisper, Seamless, OWSM, and Canary-1B. 
However, the reported data and compute requirements for their training are prohibitive for many in the research community.
In this work, we focus on the efficiency angle and ask the questions of whether we are training these speech models efficiently, and what can we do to improve?

We argue that a major, if not the most severe, detrimental factor for training efficiency is related to the sampling strategy of sequential data.
We show that negligence in mini-batch sampling leads to more than 50\% computation being spent on padding.
To that end, we study, profile, and optimize Canary-1B training to show gradual improvement in GPU utilization leading up to 5x increase in average batch sizes versus its original training settings. 
This in turn allows us to train an equivalent model using 4x less GPUs in the same wall time, or leverage the original resources and train it in 2x shorter wall time. 

Finally, we observe that the major inference bottleneck lies in the autoregressive decoder steps. 
We find that adjusting the model architecture to transfer model parameters from the decoder to the encoder results in a 3x inference speedup as measured by inverse real-time factor (RTFx) while preserving the accuracy and compute requirements for convergence.
The training code and models will be available as open-source.
\end{abstract}

\section{Introduction}
\label{sec:introduction}

The attention encoder-decoder (AED) architecture~\cite{chan2016listen} is a core component of many recent foundation speech models~\cite{radford2022robustspeechrecognitionlargescale,barrault2023seamless,peng2023reproducing,peng2024owsm,puvvada24_interspeech}. 
In this work, we scrutinize the training and inference efficiency aspect of AED models. 
For the scope of this study, we set two aims for the training of a model to be considered efficient: A) it leverages the available hardware to the full extent possible; B) it minimizes the amount of unnecessary computation. 
The first aim can be monitored with hardware utilization diagnostics, such as the percentage of GPU compute and memory utilization. 
However, aim B is defined broadly and must be further specified to be measurable. 

Speech modeling is inherently a sequence processing problem, where training examples are sequences of variable length.
Due to the design of modern deep learning frameworks such as PyTorch, individual examples must be brought to the same shape before they can form a mini-batch tensor\footnote{
It is possible to write specialized GPU kernels for processing variable-shaped batches, e.g. in~\citet{k2fsa} or~\citet{flashattn}. Such implementations are specialized for specific operations only, and as such are beyond the scope of this work.}. 
The simplest way of accommodating this constraint is to pad the examples and use appropriate padding masks in model computation to ignore padding data contribution. 
However, padding still contributes to computation which is wasteful.
Therefore, the amount of padding may be used as a proxy measure for one aspect of training efficiency.

\begin{figure}[!t]
    \centering
    \includegraphics[width=1.0\linewidth]{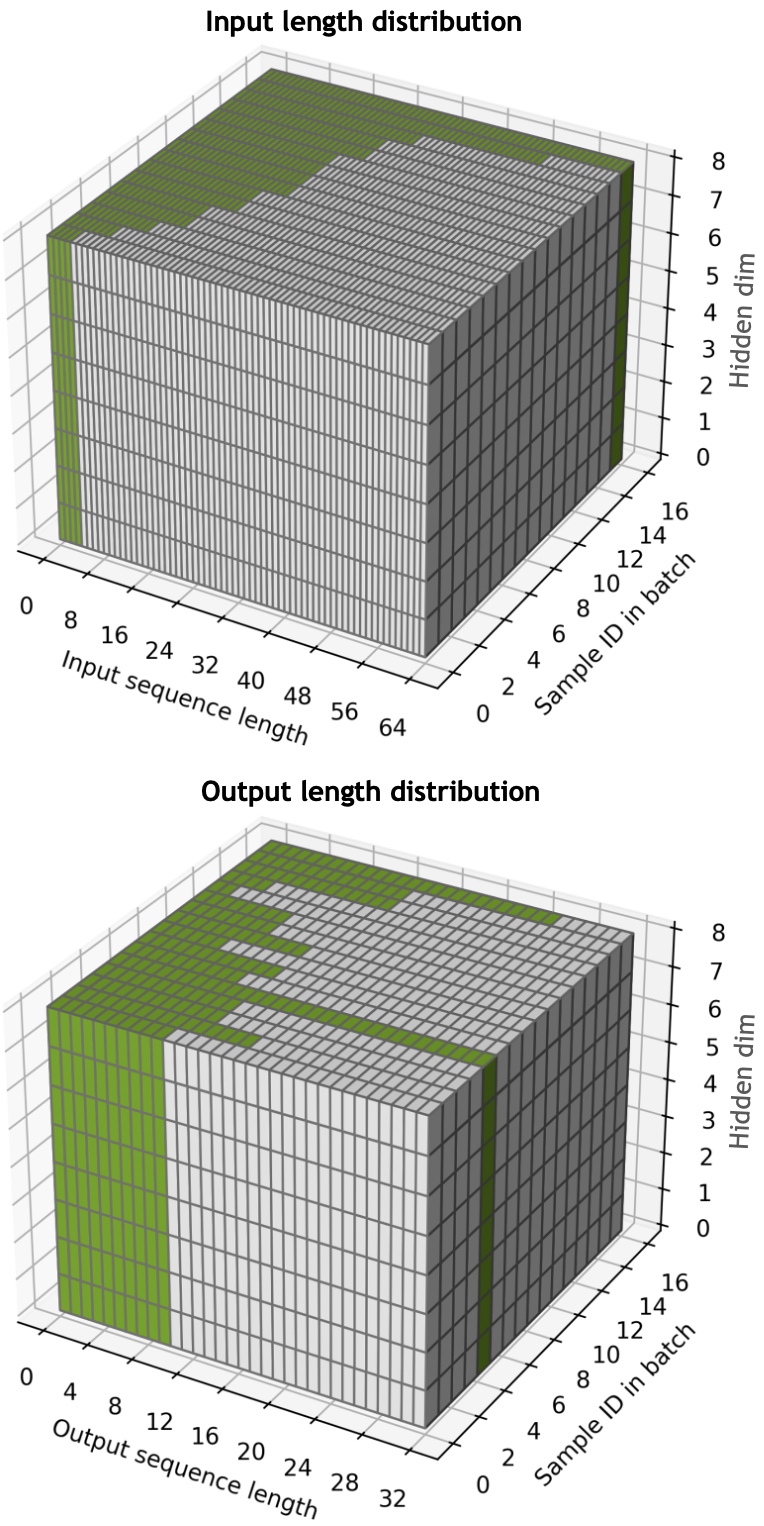}
    \caption{Visualization of a randomly sampled mini-batch representing variable length input speech and output transcription data as 3D activation tensors with shape \texttt{(batch, length, hidden\_dim)}. The sequence lengths were sampled from our training data distribution and the hidden dimension was set to 8 for readability. Grey elements indicate padding elements. There are two axes of padding, one in each tensor, limiting the efficiency of both encoder and decoder modules.}
    \label{fig:efficiency_cubes}
\end{figure}

Bucketing~\cite{7583516,doetsch2017comprehensivestudybatchconstruction} is a stratified sampling technique that populates mini-batches with examples of similar length to minimize the padding. 
However, as observed by~\citet{zelasko2024emmett}, bucketing only stratifies on a single sequence length dimension, e.g., utterance duration. 
This is not sufficient for some speech modeling tasks, such as speech recognition and translation, which have two distinct sequence dimensions: the input sequence (i.e., utterance duration) and the output sequence (i.e., the transcription). 
We illustrate this in Figure~\ref{fig:efficiency_cubes}, where there are two separate axes of padding that affect different operations in a neural network model.
\citet{zelasko2024emmett} proposes a 2D bucketing scheme to minimize the padding in both dimensions.
We study these enhancements in the context of training a state-of-the-art speech recognition and translation model, Canary-1B~\cite{puvvada24_interspeech}, and further refine them. 

Our contributions are as follows:
\begin{enumerate}
    \item We use Canary-1B as a baseline training experiment, and apply 2D bucketing and batch size optimizer originally proposed by~\citet{zelasko2024emmett} in the context of machine translation.
    \item We identify three issues with bucketing: tail-worker effect in distributed training, token-per-second outliers, and training start overhead from dynamic bucketing buffering, and propose adequate solutions.
    \item We profile the inference of Canary-1B and tune its architecture to achieve 3x faster inference without loss of accuracy.
    \item We show that as a result of all applied optimizations, Canary-1B trains using 4x less GPUs in the same wall time.
    \item Compared to fixed batch size training, our fully optimized setup converges 2x faster.
    \item The training code will be released as  open-source software.
\end{enumerate}

\section{Methods}
\label{sec:methods}

\textbf{Distributed training and synchronized bucketing.} In a distributed data parallel (DDP) training setup, each rank (typically corresponding to a single GPU) is expected to sample a different mini-batch for a given training step. This is easily achieved by seeding the RNG differently in each rank's training process. However, when combined with dynamic bucketing, this leads to bucket selection being unsynchronized between ranks. This means that one rank may draw a large batch size of short utterances, while another draws a small batch size of long utterances. Due to the model's super-linear time and/or memory complexity in the training step w.r.t. sequence lengths, this causes a tail-worker effect to appear, i.e. all (but one) ranks in DDP training have to wait for the slowest rank to finish its training step in order to globally reduce the gradients. 

We amend this issue by maintaining a separate RNG for bucket selection with a shared seed across all ranks, with one caveat. Dynamic bucketing is not guaranteed to have a mini-batch available in every bucket at any given training step due to limited buffer size. When a rank cannot sample mini-batch from a globally selected bucket, we instead select the closest bucket that has at least one mini-batch available. Note that this synchronized bucketing implementation is very efficient as it does not introduce any inter-process synchronization.

\begin{figure}
    \centering
    \includegraphics[width=1.0\linewidth]{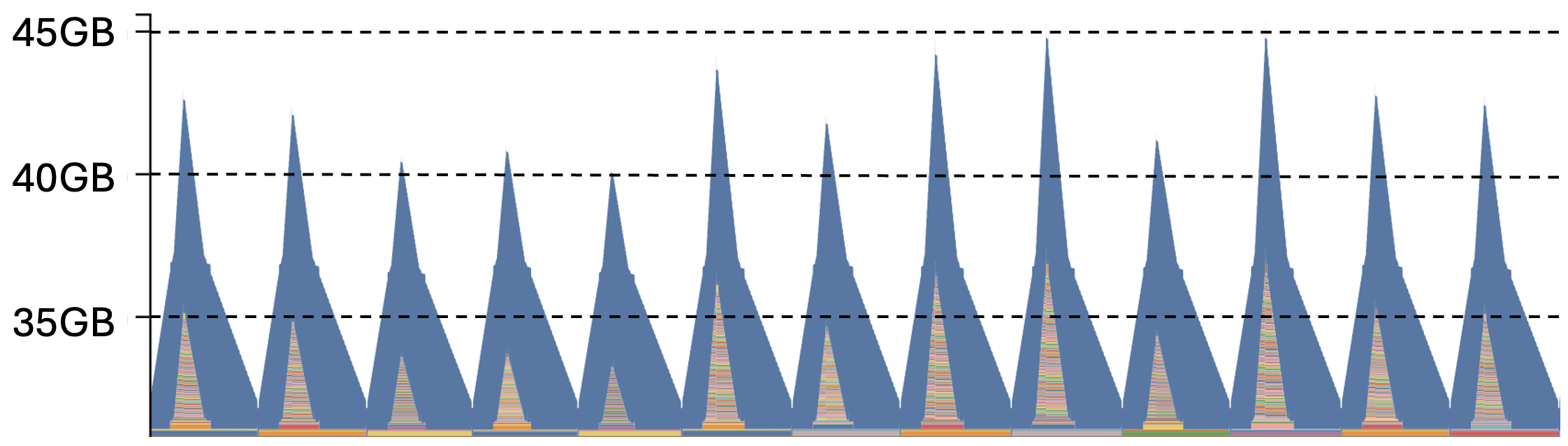}
    \caption{Memory usage profile of Canary-1B training on RTX 6000 Ada 48GB GPU using 1D dynamic bucketing with equal batch duration heuristic. Each peak denotes the point right after training loss computation for a single training step. The memory usage for majority of training steps is well below the maximum, showing room for efficiency improvement.}
    \label{fig:memprof}
\end{figure}

\begin{figure}
    \centering
    \includegraphics[width=1.0\linewidth]{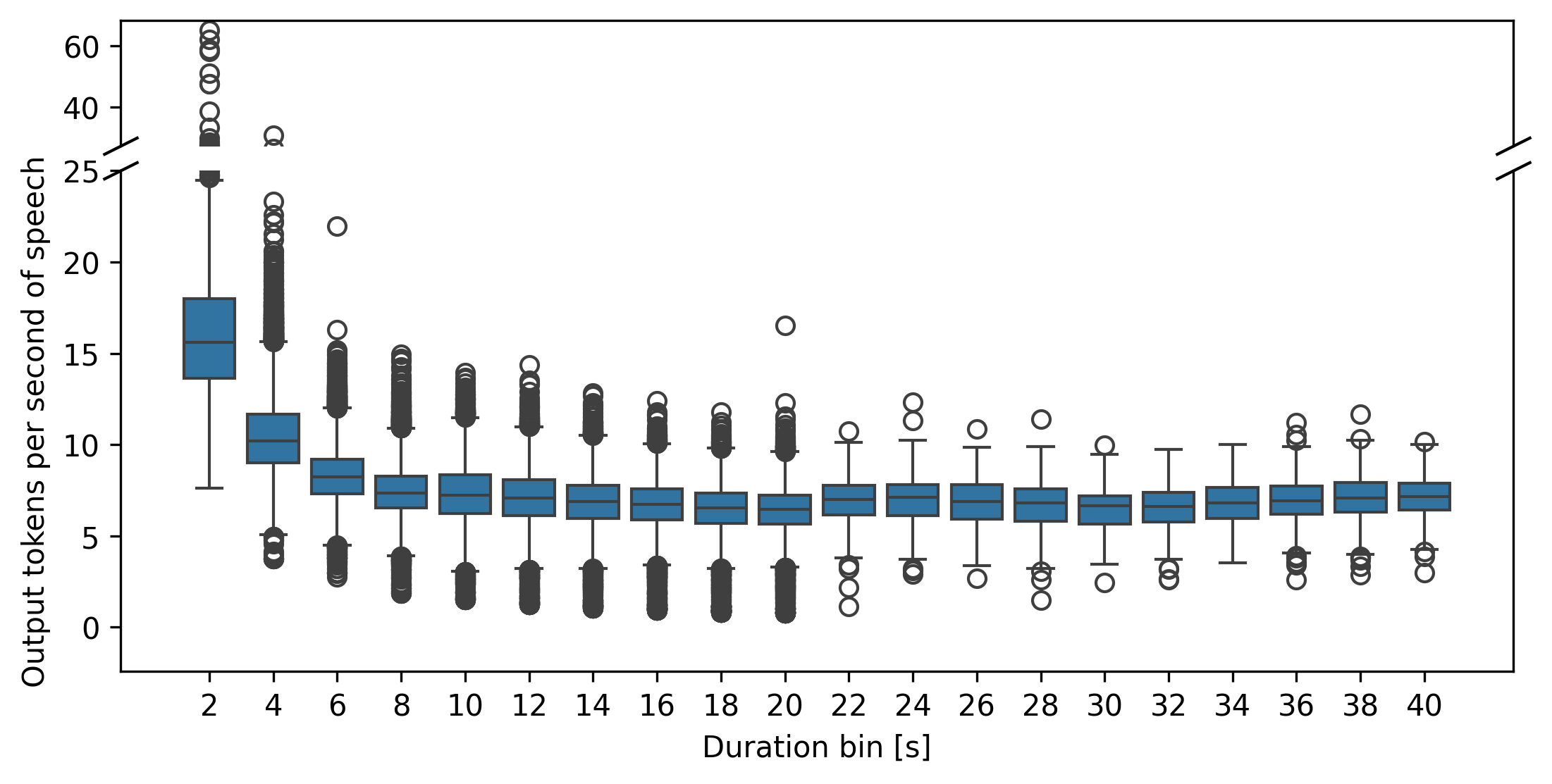}
    \caption{Output token rate distribution on a 100k sample of Canary-1B-Flash training data. Utterances are grouped into duration bins with 2s increment. Short utterances have significantly more transcript tokens per second, partially due to a fixed-length prompt fed to the decoder. This highlights the need for careful data filtering and tuning of 2D bucketing settings.}
    \label{fig:tpsdist}
\end{figure}

\textbf{Output token rate distribution and token-per-second (TPS) filtering.} 
Originally, Lhotse~\cite{zelasko2021lhotse} dynamic bucketing used a cumulative batch duration heuristic to determine the batch size for each bucket dynamically.
The sampler would keep drawing examples until the cumulative speech duration in a batch exceeds a set threshold, naturally leading to smaller batch sizes for longer utterances. 
On the surface, this approach appears efficient, often indicating that the allocated GPU memory is close to the maximum for the entire training.
However, through a closer inspection with PyTorch memory profiler in Figure~\ref{fig:memprof}, we observed that this heuristic leads to an unpredictable GPU memory usage pattern, making it difficult to tune the threshold well and to ensure full GPU utilization. 
Upon closer study, we discovered that the training sometimes runs into out-of-memory issues on mini-batches with output sequence length outliers--i.e., examples with unusually long transcripts. 
Such high variance of GPU memory usage for mini-batches drawn from the same bucket highlights the importance of additional stratification on output sequence length in 2D bucketing. 
We show an example TPS distribution in Figure~\ref{fig:tpsdist}.

The right solution to this issue requires a closer look at the training data. In some cases, high TPS outliers correspond to low-quality data, such as hallucinations from synthetic data generation stage. In such event, it is appropriate to incorporate a TPS filter before bucketing sampler. However, in other scenarios high TPS data may be very well expected. Examples include augmenting transcripts with word-level timestamps or multilingual training with languages with vastly different token count distributions. 

To this end, we augment the 2D bucketing method proposed by~\citet{zelasko2024emmett} with a modified sample-to-bucket allocation algorithm. We consider the baseline 2D bucketing algorithm as \textit{strict}: to match a sample to a bucket, it first determines a 1D bucket based on the sample's input sequence length, and then chooses one of its sub-buckets based on the sample's output sequence length. If the sample has longer transcript than what was found as the upper bound during bucket bin estimation, it would be discarded. In our modified \textit{flexible} approach, we instead search for the smallest bucket that can fit a given sample. That means we may allocate that sample to a bucket corresponding to longer utterances with longer transcipts, sacrificing some input padding for the ability to keep the outlier sample.

\textbf{Concurrent bucketing.}
Dynamic bucketing, as implemented in Lhotse, maintains a fixed size in-memory buffer for training examples to partition them into buckets. 
With sequential IO formats such as webdataset~\cite{webdataset} or Lhotse Shar~\cite{zelasko2021lhotse} this means reading a number of audio recordings into memory, which in our setup resulted in a 5-10 minute overhead at the start of training. 
Since in practice a single training run is composed of many time-limited scheduler jobs, this overhead becomes significant at scale.
We extended the dynamic bucketing sampler with a producer thread that reads examples and populates a thread-safe queue. 
We adjust the existing consumer thread to wait until the queue is 10\% populated and then start sampling mini-batches, reducing the overhead to below one minute.
We find that in practice, data reading is much faster than the training step, which lets the producer fill the queue entirely after several more minutes while the training is already ongoing.
Note that simply decreasing the buffer size impacts the randomness of sampling.

\textbf{Transferring model capacity from decoder to encoder.} 
Upon model's inference profiling, we noticed that the majority of the computation time is taken by the autoregressive cross-attention decoder. 
Its impact can be easily visualized by an order-of-magnitude gap between real-time factors (RTFx) of attention-encoder-decoder models (RTFx=235), and an otherwise similar 1B CTC~\cite{graves2006connectionist} encoder-only model (RTFx=2728) available in HuggingFace Open ASR Leaderboard\footnote{\url{https://huggingface.co/spaces/hf-audio/open_asr_leaderboard}}. 
Similarly as Whisper-v3-turbo~\cite{radford2022robustspeechrecognitionlargescale}, we decrease the number of Canary-1B's decoder layers from 24 to 4. 
However, this change reduces the model's parameter count from 1016M to 680M, making it almost twice smaller. 
As a result we observed degraded prediction accuracy, especially for translation. 
This is consistent with Whisper-v3-turbo findings. 
Further, we find that this degradation can be completely compensated by increasing the encoder's capacity with minimal impact on inference speed, which we demonstrate in Section~\ref{sec:results}. 
Our final configuration increases the number of encoder layers from 24 to 32 with a total of 882M parameters and is further referred to as Canary-1B-Flash.

\begin{table}[]
    \centering
    \caption{Training efficiency gains from synchronized bucketing implementation. The gain grows with distributed training's size due to increased severity of the tail-worker effect when bucketing is not synchronized.}
    \begin{tabular}{|c|c|}
        \hline
        GPUs & Training step speedup [\%] \\
        \hline
        2 & 7 \\
        16 & 13 \\
        128 & 20 \\
        \hline
    \end{tabular}
    \label{tab:syncbuckets}
\end{table}

\section{Experimental setup}
\label{sec:setup}

\begin{figure*}[!htb]
    \centering
    \includegraphics[width=1.0\linewidth]{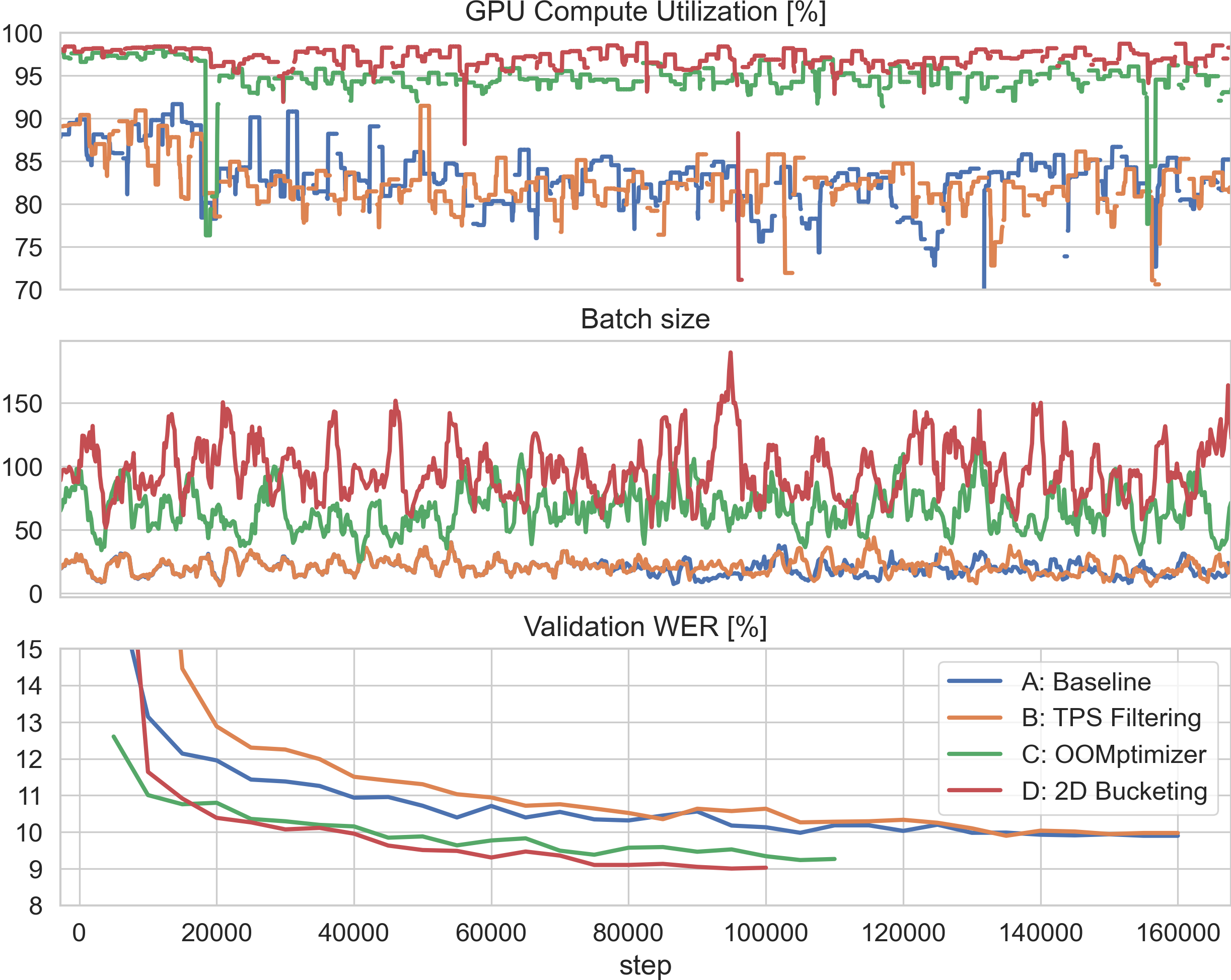}
    \caption{Canary-1B training efficiency comparison across four training schemes. Scheme A is Canary-1B baseline. Scheme B adds TPS filtering, freeing up GPU memory. Scheme C replaces batch duration heuristic with OOMptimizer for batch size estimation. Scheme D adds 2D bucketing to further reduce the number of padding tokens. The efficiency gains directly translate to quicker validation WER convergence. The horizontal axes for all metrics except for WER demonstrate the first 100k training steps.}
    \label{fig:bigfig}
\end{figure*}

Unless otherwise stated, we adopt the same data, training, and evaluation setup as in~\citet{puvvada24_interspeech}, resulting in a training set of 85k hours of speech recognition data in English, French, German, and Spanish, complemented by synthetically generated translations for the speech translation task\footnote{Megatron NMT model used: \url{https://catalog.ngc.nvidia.com/orgs/nvidia/teams/nemo/models/megatronnmt_any_en_500m}}. 
For our baseline we also adopt the model architecture and training hyperparameters from~\citet{puvvada24_interspeech}, meaning every model in this work has its encoder initialized from a pretrained ASR checkpoint\footnote{\url{https://catalog.ngc.nvidia.com/orgs/nvidia/teams/nemo/models/stt_multilingual_fastconformer_hybrid_large_pc_blend_eu}}.
When we increase the number of encoder layers, the additional top layers are initialized randomly, and the decoder is always initialized randomly.
The models are trained on NVIDIA A100 80GB GPUs.

We evaluate speech recognition with word error rate (WER) on Open ASR Leaderboard and speech translation with COMET~\cite{rei-etal-2020-comet} scores on FLEURS~\cite{fleurs2022arxiv} and CoVOST v2~\cite{wang21s_interspeech} datasets. For COMET metric we used \texttt{Unbabel/wmt22-comet-da} model with `unbabel-comet` version 2.2.2. For tracking validation metrics, we use BLEU scores computed using SacreBLEU library~\cite{post2018call}.
The validation sets for each language are taken from Mozilla CommonVoice 12~\cite{commonvoice:2020} for speech recognition and from FLEURS and CoVOST v2 for speech translation.

\begin{table}[]
    \centering
    \caption{Canary-1B and Canary-1B-Flash final evaluation WER on HuggingFace Open ASR Leaderboard as a function of number of GPUs and wall time spent on training. Optimizations include TPS filtering, 2D bucketing, synchronized bucketing, and OOMptimizer. Canary-1B-Flash is trained only with efficiency optimizations.}
    \begin{tabular}{|l|c|c|c|}
        \hline
        Experiment & GPUs & Runtime & WER [\%] \\
        \hline
        Canary-1B & 128 & 36h & 6.54 \\
        \quad\textit{+optimized} & 32 & 36h & 6.51 \\
         & 128 & 19h & 6.47 \\
        \hline
        1B-Flash & 32 & 38h & 6.5 \\
         & 128 & 46h & 6.35 \\
        \hline
    \end{tabular}
    \label{tab:testresults}
\end{table}

\begin{table}[]
    \centering
    \caption{RTFx comparison between baseline Canary-1B and the modified architectures explored in this work. This number is interpreted as how many seconds of recorded speech can be transcribed in one second of wall time. These values were measured on a single RTX 6000 Ada 48GB GPU.}
    \begin{tabular}{|l|c|c|}
        \hline
        Model & Parameter count & RTFx \\
        \hline
        Canary-1B & 1018M & 345 \\
        \textit{+ small decoder} & 680M & 1097 \\
        \textit{+ larger encoder} & 882M & 992 \\
        \hline
    \end{tabular}
    \label{tab:rtfx}
\end{table}

\textbf{Training data sampling.} 
For baseline Canary-1B, we use 30 buckets ranging from 0.5s to 40s with bins estimated for equal occupancy w.r.t. cumulative bin duration on a 100k sample of training examples. 
The batch sizes for each bucket are determined using 360s cumulative batch duration threshold mentioned in Section~\ref{sec:methods}.
The 2D bucketing setup leverages a 30x2 bucket configuration following~\citet{zelasko2024emmett}, which means that each of 30 duration bins is further sub-divided into 2 token bins. 
Unless otherwise indicated, we calibrate the batch sizes for each bucket with a batch size optimizer~\cite{zelasko2024emmett}. 
All experiments except for the baseline use a TPS filter set at 25. 
During early experiments, we noticed that Canary-1B-Flash has convergence stability issues due to outliers above that threshold. 
A closer inspection revealed this is due to low quality synthetic translation examples from Canary-1B training set. 
This issue was not originally noticed by~\citet{puvvada24_interspeech}, where the model used 24 decoder layers.
It may be indicative that the training of a model with a smaller decoder is less resilient against inaccurate labels.

\textbf{Experiments.} We present the following experiments to validate our claims:
\begin{itemize}
    \item[\textbullet] [E1] Canary-1B training with and without synchronized bucketing. The improvement is measured by relative training step time reduction.
    \item[\textbullet] [E2] Canary-1B training with different data sampling schemes. The improvement is illustrated using multiple metrics including GPU compute and memory utilization, amount of padding, and convergence speed:
    \begin{itemize}
        \item[A.] Baseline from~\citet{puvvada24_interspeech} with batch duration heuristic.
        \item[B.] Same as A, but with TPS filtering set at 25 tokens-per-second.
        \item[C.] Same as B, but with batch size optimizer replacing the heuristic.
        \item[D.] Same as C, but with 2D bucketing (30x2) replacing 1D bucketing (30). Bucket batch sizes are re-estimated.
    \end{itemize}
    \item[\textbullet] [E3] Canary-1B-Flash architectural changes ablations, where Canary-1B serves as the baseline. We measure WER, COMET, and inference speed.
    \item[\textbullet] [E4] Canary-1B-Flash convergence speed comparison between the widely adopted fixed batch size strategy and proposed fully optimized setup.
\end{itemize}
Finally, we report the total training time and resources needed to reach original Canary-1B's performance level for both Canary-1B and Canary-1B-Flash training with the full set of introduced optimizations.

\section{Results}
\label{sec:results}

\textbf{Synchronized bucketing [E1].} 
We measure the mean time needed to execute 1000 training steps for Canary-1B in three distributed settings: with 2, 16, and 128 GPUs. 
When we turn on synchronized bucketing, we observe a speedup starting from 7\% for 2 GPUs and growing to 20\% for 128 GPUs in Table~\ref{tab:syncbuckets}. 
The increasing efficiency gain with training size scaling is in line with our expectations outlined in Section~\ref{sec:methods}.

\textbf{2D bucketing and OOMptimizer [E2].}
We show the effect of applying TPS filtering, OOMptimizer, and 2D bucketing one-by-one in Figure~\ref{fig:bigfig}. 
With TPS filtering alone, we notice the convergence speed is initially slower, but catches up in later training stage. 
However, it partially amends the issue seen in~\ref{fig:memprof} through reducing the peak GPU memory allocation by 20\%, allowing to increase the batch size further in the next steps.
Replacing the equal batch duration heuristic with bucket batch sizes tuned by OOMptimizer increased the mean batch size by 3.4 times and mean GPU utilization by 20\%. 
Further adding 2D bucketing resulted in a total of 5x mean batch size increase compared to the baseline.

While larger batch sizes and GPU utilization are useful, the real value of these contributions is in the reduction of resources and time required to achieve an equivalent result in terms of model accuracy.
In Table~\ref{tab:testresults} we compare the resources and time required to train original Canary-1B with its fully optimized training scheme.
Row 2 demonstrates that the introduced optimizations let us to train the same model in the same amount of time (36 hours) by using 4x less GPUs. 
Row 3 shows that if we retain the same resources, we may train the model in 2x shorter time.

\begin{table*}
\centering
\caption{Ablation study for Canary-1B-Flash architecture design based on speech translation performance. We report COMET scores on FLEURS and COVOST, translating from English to German, Spanish, and French, and in the opposite direction. For readability, the COMET scores are multiplied by 100.}
\begin{tabular}{|l|lll|lll|l|lll|l|l|}
\hline
\multirow{2}{*}{Model} & \multicolumn{3}{c|}{COVOST ($\rightarrow$EN)}                                      & \multicolumn{3}{c|}{FLEURS ($\rightarrow$EN)}                                      & \multirow{2}{*}{\shortstack{X$\rightarrow$EN \\ AVG}} & \multicolumn{3}{c|}{FLEURS (EN$\rightarrow$)}                                      & \multirow{2}{*}{\shortstack{EN$\rightarrow$X \\ AVG}} & \multirow{2}{*}{AVG} \\
                       & DE & ES & FR & DE & ES & FR & & DE & ES & FR &                                         &                      \\
\hline
Canary-1B & 82.4 & 85.4 & 83.4 & 84.2 & 81.5 & 83.2 & 83.3 & 81.4 & 81.1 & 81.6 & 81.4 & 82.7 \\
\textit{+sm. dec.} & 81.2 & 85.0 & 83.3 & 83.0 & 81.4 & 83.0 & 82.8 & 80.1 & 80.8 & 80.9 & 80.6 & 82.1 \\
\textit{+lg. enc.} & 83.6 & 86.0 & 84.2 & 85.3 & 82.4 & 84.5 & 84.3 & 81.3 & 81.7 & 82.6 & 81.9 & 83.5 \\
\hline
\end{tabular}
\label{tab:tiarchresults}
\end{table*}

\textbf{Canary-1B-Flash architectural changes [E3].}
First, we show the effect of Canary-1B-Flash architecture on inference speed in Table~\ref{tab:rtfx}. 
Decreasing the decoder size yields a major 3.2x improvement in RTFx, and the increase in encoder size manages to retain 2.9x improvment in RTFx compared to the baseline.
Note that despite inference speedups, the training step speed is roughly the same for all variants, because we leverage the efficiency gains to further increase the batch size (tuned again with OOMptimizer).

Table~\ref{tab:tiarchresults} shows the translation quality results of Canary-1B and Canary-1B-Flash measured by COMET score. 
As we mentioned in Section~\ref{sec:methods}, decreasing decoder size primarily impacts translation, but these losses fully recovered by transferring the capacity to decoder--in fact, the Canary-1B-Flash model outperforms the baseline in this setting. 
Our findings are consistent with~\citet{kasai2021deep}.

We present the ASR results for fully trained optimized Canary-1B-Flash in rows 4 and 5 in Table~\ref{tab:testresults}. 
We see that the overall ASR performance is retained with the same amount of compute as used for optimized Canary-1B. 
When using the original amount of resources and training for 25\% longer, we further achieve a state-of-the-art WER of 6.35\% on the Open ASR Leaderboard.

\begin{figure}
    \centering
    \includegraphics[width=1.0\linewidth]{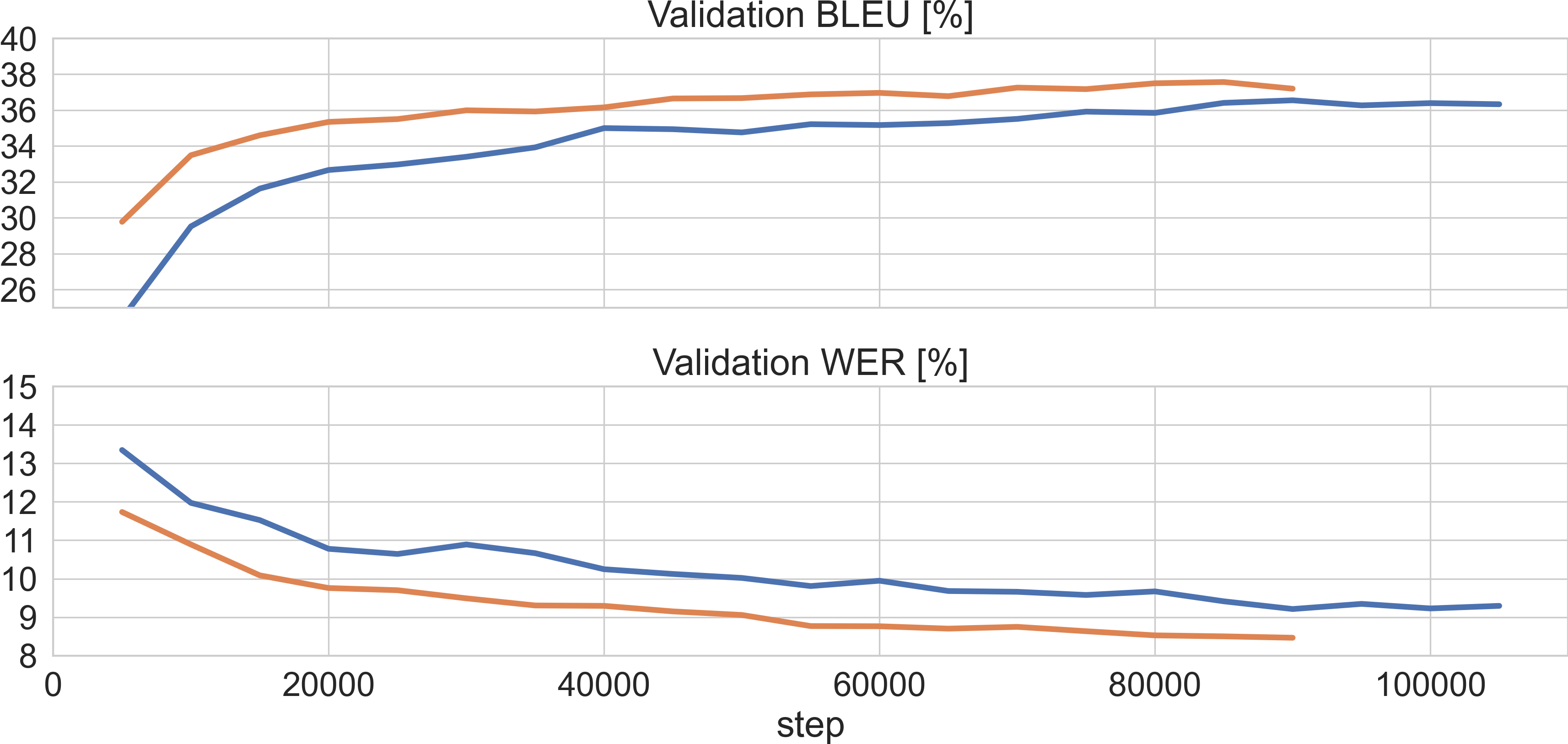}
    \caption{Comparison of Canary-1B-Flash convergence speed with fully optimized 2D bucketing scheme (orange) vs fixed batch size of 768 (blue), both on 32 GPUs.}
    \label{fig:whisperfig}
\end{figure}

\textbf{Convergence speed advantages versus fixed batch size training [E4].} 
Finally, the reader might be tempted to ask: does the proposed method indeed improve compared to simply training with a fixed batch size, similarly to Whisper~\cite{radford2022robustspeechrecognitionlargescale} or OWSM~\cite{peng2024owsm}?
We answer this question by presenting validation WER and BLEU plots for both training schemes in Figure~\ref{fig:whisperfig}.
The same WER or BLEU values are achieved with roughly 2x more training steps by the fixed batch size scheme, with training step time being approximately the same.
In our setup the fixed batch size scheme requires padding every mini-batch to 40 seconds duration, resulting in 57\% padding of audio and 59\% padding of transcripts on average.
Notably it maintains high GPU compute and memory utilization, but they are mostly spent on computing padding.
For comparison, with 2D bucketing in a 30x2 configuration we achieved as little as 4.5\% padding of audio and 19\% padding of the transcripts.
We noticed that driving the transcript padding ratio lower is difficult due to the fact that longer recordings may contain little or no speech, resulting in a wider spread of output sequence lengths.
Increasing the number of 2D buckets further did not yield meaningful improvement in this setup.

\section{Related work}
\label{sec:related_work}

Whisper~\cite{radford2022robustspeechrecognitionlargescale} is a transformer~\cite{vaswani2017attention} attention encoder-decoder (AED) model~\cite{bahdanau2015neural} that has demonstrated impressive ASR and AST capabilities in 96 languages. 
It was initially trained with 680K hours of data (v1 and v2) and later extended to 5M hours (v3), out of which 4M were transcribed by an earlier model version. 
In DistilWhisper, \citet{gandhi2023distilwhisperrobustknowledgedistillation} optimized Whisper's architecture for inference, noticing that distilled model still works well when only as little as two decoder layers are retained.
As a result, Whisper was also released in a \texttt{turbo} variant that decreased the number of decoder layers to 4 and was fine-tuned from a larger initial model. Notably, Whisper \texttt{turbo} was fine-tuned exclusively on speech recognition data, as the authors claimed they did not expect the model to perform well on translation\footnote{\url{https://github.com/openai/whisper/discussions/2363}}. 

Seamless~\cite{barrault2023seamless} is a multimodal streaming translation model supporting around 100 languages. It uses several components pretrained on over 4M unlabeled hours of speech, which are later fine-tuned jointly on 125k hours.

OWSM~\cite{peng2023reproducing} is the first fully open-source attempt at reproducing  Whisper model. It was trained on 180k hours of publicly available data and supports 151 languages. OWSM v3.1 adopted E-Branchformer architecture, achieving superior accuracy and speed~\cite{peng2024owsm}.

Canary-1B~\cite{puvvada24_interspeech} is an attention encoder-decoder model trained for speech recognition and translation with punctuation and capitalization recovery. It uses a FastConformer~\cite{rekesh2023fast} encoder architecture that is initialized from a pretrained RNN-T~\cite{graves2012sequencetransductionrecurrentneural} ASR model for quicker convergence. 
Canary-1B achieved similar level of translation performance to Whisper and Seamless despite being trained on less data, and using exclusively synthetic translation data.

EMMeTT~\cite{zelasko2024emmett} introduced the concepts of 2D bucketing and batch size optimizer--OOMptimizer--to accelerate the training of large language models (LLM) extended with a speech encoder for multimodal machine translation capability. 
OOMptimizer algorithm is a variant of bisection that simulates model training steps on artificial data of various shapes to determine the maximal batch size for each of sequence length buckets.
This extra tuning step is performed before model training.

\section{Conclusion}
\label{sec:conclusion}

We demonstrated that stratified sampling is critical to achieve efficient training of attention-encoder-decoder speech models on the example of Canary-1B. 
We refined the methods proposed by~\citet{zelasko2021lhotse,puvvada24_interspeech,zelasko2024emmett} to achieve a 4x reduction in GPU resources to train an equivalent model, without writing specialized kernels for any of the operations. 
We also showed that equivalently, the same amount of compute may be used to train the model in 2x less time.
Further, we optimized the model for 3x faster inference without loss of accuracy through shifting the parameters from the decoder to the encoder.
Perhaps somewhat surprisingly, Canary-1B-Flash is able to effectively learn speech translation despite having a smaller decoder, as we found that transferring the parameter budget to the encoder both prevents accuracy loss and has a marginal impact on the speed of inference.
We emphasize that our proposed training optimization method did not require changing of a single line of code of the training script or the model's logic--it is sufficient to adapt the data sampling module.

Although this work focuses on AED models of speech, a similar analysis of the sequence length and output token rate distributions may be leveraged for sampling stratification and increased training efficiency in other sequence-to-sequence modeling problems.
The training code, together with Canary-1B-Flash model will be available as open-source.

\section*{Limitations}

This work studies the training and inference efficiency of models sized at between 600M and 1B parameters with a relatively large training dataset of 85k hours of speech. 
The main efficiency gains stem from the ability to increase the average batch size in training, which may or may not be applicable to smaller dataset and/or model setups characterized by a lower critical batch size~\cite{mccandlish2018empiricalmodellargebatchtraining,shallue2019measuring,zhang2024doescriticalbatchsize}. 
Conversely, models of larger size typically require some form of model parallelism for their training, which may require significant adjustments in the training setup to accommodate dynamically shaped batches, or to estimate the bucket batch sizes with OOMptimizer algorithm.

The models studied in this work are trained on four languages (English, French, Spanish, and German) which may be considered as high-resource. 
The considered baseline models are of similar size and architecture, but extend their support to about 100 languages. 
Parts of the data used to train our models are not publicly available.
We used training data with no personally identifiable information or offensive content.
The model is trained to translate from English to any of the other languages and vice versa, but not between a pair of non-English languages.
For speech translation task, the model is trained entirely on synthetic data, and is likely to carry over the biases and errors present in the machine translation model used for generation.
The synthetic data was only lightly filtered to discard severe hallucinations according to output token rate thresholds, as explained in Section~\ref{sec:setup}.

While this work is focused primarily on training efficiency methods, we release one of the trained models, Canary-1B-Flash, that is intended for speech recognition and translation in English, German, French, and Spanish.
Given the ubiquity of ASR and translation technology in these languages, we don't believe the model introduces any novel risks.
We will indicate the specific license for the release at the time the paper is camera-ready.

\section*{Ethical considerations}

As outlined in~\citet{hazirbas2021towards}, we assessed the Canary-1B-Flash model from row 5 of Table~\ref{tab:testresults} for age and gender bias using the Casual Conversations v1 dataset. 
The results are presented in Table~\ref{tab:gender_bias} and Table~\ref{tab:age_bias}. 
This evaluation is limited to English speech recognition.

\begin{table}
    \centering
    \begin{tabular}{|l|l|l|l|l|}
        \hline
        Gender & Male & Female & N/A & Other \\
        \hline
        Count & 19325 & 24532 & 926 & 33 \\
        WER [\%] & 14.66 & 12.44 & 17.17 & 27.56 \\
        \hline
    \end{tabular}
    \caption{The results of Canary-1B-Flash's evaluation for gender bias in English speech recognition.}
    \label{tab:gender_bias}
\end{table}

\begin{table}
    \centering
    \begin{tabular}{|l|l|l|l|l|}
        \hline
        Age & 18-30 & 31-45 & 46-85 & 1-100 \\
        \hline
        Count & 15956 & 14585 & 13349 & 43890 \\
        WER [\%] & 13.18 & 13.45 & 13.64 & 13.41 \\
        \hline
    \end{tabular}
    \caption{The results of Canary-1B-Flash's evaluation for age bias in English speech recognition.}
    \label{tab:age_bias}
\end{table}

\section*{Acknowledgments}

The authors thank Travis Bartley and Somshubra Majumdar for insightful discussions about the model architecture modifications.

\bibliography{custom}

\appendix

\clearpage
\section{Appendix A: Speech translation evaluation with BLEU scores}
\label{sec:appendix_a}

\begin{table*}[t]
\centering
\caption{Ablation study for Canary-1B-Flash architecture design based on speech translation performance. We report SacreBLEU scores on FLEURS and COVOST, translating from English to German, Spanish, and French, and in the opposite direction. We include comparison with other baseline models that report BLEU scores.}
\begin{tabular}{|l|lll|lll|lll|}
\hline
\multirow{2}{*}{Model} & \multicolumn{3}{c|}{COVOST ($\rightarrow$EN)}                                      & \multicolumn{3}{c|}{FLEURS ($\rightarrow$EN)} & \multicolumn{3}{c|}{FLEURS (EN$\rightarrow$)} \\
                       & DE & ES & FR & DE & ES & FR & DE & ES & FR \\
\hline
OWSM-v3.1 (1B) & 18.1 & 23.9 & 24.5 & 13.2 & 9.4 & 12.4 & 24.4 & 11.4 & 16.4 \\
Whisper-large-v3 (1.5B) & 34.2 & 39.2 & 35.5 & 33.4 & 22.7 & 31.0 & - & - & - \\
SeamlessM4T-medium (1.2B) & 35.6 & 39.2 & 39.3 & 33.4 & 21.7 & 30.9 & 28.3 & 21.1 & 37.4 \\
SeamlessM4T-large-v2 (2.3B) & 40.0 & 42.9 & 42.1 & 37.1 & 25.4 & 33.7 & 33.2 & 23.7 & 43.1 \\
\hline
Canary-1B & 37.0 & 40.3 & 40.0 & 32.7 & 22.0 & 31.1 & 31.4 & 22.4 & 40.2 \\
\textit{+sm. dec.} (680M) & 35.9 & 40.2 & 39.7 & 32.1 & 21.3 & 30.7 & 29.8 & 21.7 & 38.5 \\
\textit{+lg. enc.} (880M) & 37.9 & 40.7 & 40.4 & 34.5 & 23.0 & 32.1 & 32.5 & 22.4 & 40.0 \\
\hline
\end{tabular}
\label{tab:tiarchresults_bleu}
\end{table*}

We report the speech translation evaluation results in BLEU scores in Table~\ref{tab:tiarchresults_bleu} for an easier comparison with other foundation speech models that did not report COMET scores. The patterns observed we observed with COMET evaluation in Table~\ref{tab:tiarchresults} hold. The optimized Canary-1B-Flash model maintains the advantage over other similarly sized models.
Given that the machine translation community has found COMET to be more reliable~\cite{mathur-etal-2020-tangled}, we encourage the readers to consult Table~\ref{tab:tiarchresults}.

\end{document}